\documentclass[11pt]{article}

\usepackage[preprint]{acl}

\usepackage{times}
\usepackage{latexsym}

\usepackage[T1]{fontenc}

\usepackage[utf8]{inputenc}

\usepackage{microtype}

\usepackage{inconsolata}

\usepackage{color, colortbl}
\definecolor{Gray}{gray}{0.9}
\usepackage{multirow,adjustbox}
\usepackage{arydshln}
\usepackage{diagbox}
\usepackage{subcaption}
\usepackage{todonotes}
\usepackage[dvipsnames]{xcolor}
\usepackage{amsmath, amsfonts}
\usepackage[most]{tcolorbox}
\newtcolorbox{promptbox}[2][]{%
  breakable,
  colback=gray!5,
  colframe=gray!80,
  fonttitle=\bfseries,
  title={#2},   
  #1             
}
\usepackage{bbold}

\usepackage{bbm}

\usepackage{threeparttable}

\title{Understanding the Effects of Domain Finetuning on LLMs}

\author{
 Eshaan Tanwar\textsuperscript{1}\quad
 Deepak Nathani\textsuperscript{2}\quad
 \textbf{William Yang Wang\textsuperscript{2}}\quad
 \textbf{Tanmoy Chakraborty\textsuperscript{3}}\\[0.5em] 
 \textsuperscript{1} University of Copenhagen\quad
 \textsuperscript{2}University of California, Santa Barbara\\
 \textsuperscript{3}Indian Institute of Technology, Delhi\quad\\[0.25em]
 \small{\texttt{eshaantanwar2000@gmail.com}}
}

\begin{document}
\maketitle
\begin{abstract}
Large Language Models (LLMs) fine-tuned for specific domains exhibit strong performance; however, the underlying mechanisms by which this fine-tuning reshapes their parametric space are not well understood. Prior works primarily focus on auto-regressive or general-purpose instruct models, leaving domain-specialised LLMs under-explored. We present the first systematic study of domain-specific fine-tuning in large medical language models. Our analysis reveals that fine-tuning modifies only a small subset of the representational subspace, essentially preserving the pre-trained model’s representation. To interpret these changes in subspaces, we propose \textit{tuning vectors}, a novel framework inspired by task vectors, which explicitly capture the directional parameter shifts induced by fine-tuning. We demonstrate that these vectors are critical for enhancing both instruction-following and generation quality. Furthermore, combining tuning vectors across different domains yields improved generalisation. Upon closer inspection of directional alignment, we find these vectors primarily write new directional information into the MLP layers of the model, while amplifying existing directions in attention heads.  Our findings offer new insights into LLM adaptation and provide a general, interpretable framework for analysing specialisation in large language models.
\end{abstract}

\section{Introduction}

Foundational language models have demonstrated remarkable generalisation and understanding across a wide range of  tasks~\citep{10.5555/3495724.3495883,tai-etal-2020-exbert}. This success has led to extensive research on adapting these models to specific domains, such as legal, medical, or finance, through fine-tuning or continued pretraining. Notably, domain-adapted models often achieve strong performance with relatively small amounts of domain data~\citep{houlsby2019parameter,garcia2025aloe,hui2024qwen2,yang2024qwen25mathtechnicalreportmathematical}, highlighting their efficiency and transfer capabilities. Hence, understanding how domain-specific tuning reshapes these models is crucial for understanding the internal changes brought by these techniques. Such understanding is particularly important for building trustworthy and reliable systems in sensitive domains~\citep{kumar2024trustworthiness}.

Prior research has sought to uncover how finetuning affects the representation space of foundational models, revealing that domain adaptation induces structured changes in model parameters and activations. 
\citet{gururangan-etal-2020-dont} demonstrated that continued pretraining on domain-specific corpora improves downstream performance by altering activation patterns. Similarly, \citet{merchant-etal-2020-happens} and \citet{zhou-srikumar-2022-closer} showed that such changes preserve core linguistic features while enabling the model to learn new features correlated with specific tasks. \citet{ilharco2022editing} further introduced task vectors to capture how finetuning moves model parameters in a specific task-oriented direction. Collectively, these findings suggest that finetuning encodes domain-specific knowledge. However, these studies have primarily focused on auto-encoder models or smaller models and have not addressed how large language models (LLMs), which are now more widely deployed, are adapted. Further, they primarily focus on performance improvement along one axis, for example improvement in a particular task~\citet{ilharco2022editing}, unlike LLMs' fine-tuning, which improves their generation quality across multiple axes, such as instruction-following ability, tool integration, benchmark performance, etc.

Therefore, a more recent line of research, known as \textit{model diffing}, focuses on understanding the parametric differences between a pretrained base LLM and its instruction-tuned variant. These studies~\citep{minder2025robustly, ortiz2023task, zhang-etal-2023-fine} have shown that base and instruction-tuned models share a significant portion of their encoded features. \citet{lin2023unlocking} analysed the decoding mechanisms of instruction-tuned models, showing that finetuning alters their stylistic choices in response to given instructions. Similarly, \citet{wu-etal-2024-language} demonstrated that instruction tuning modifies the model’s attention patterns, causing it to focus more on the instruction component of the prompt.

While these studies uncover several important mechanisms underlying instruction-tuned LLMs, they primarily focus on general-purpose chat models. The rapidly growing ecosystem of domain-specific, specialised LLMs has not yet been systematically studied. To address this gap, we focus on medical language models as a representative use case for examining the effects of domain-specific fine-tuning. To the best of our knowledge, we are the first to conduct a systematic investigation in this direction.

Our initial experiments reveal that domain-specific finetuning alters only a small subset of the model’s representational subspaces. To interpret these changes, we introduce \textit{tuning vectors}, a novel framework inspired by the concept of task vectors. These vectors capture how fine-tuning modifies pretrained representations. We observe that the absence of these vectors leads to models that perform poorly on instruction-following tasks. Further, we utilise these tuning vectors to analyse the differences induced by fine-tuning in pre-trained models. This approach provides a generalised method for understanding the effects of fine-tuning on domain-specific models. We summarise our findings below:

\begin{enumerate}
\item Finetuning changes only a small portion of the representational subspaces. Neural activations remain largely consistent between pretrained and fine-tuned models, and their parametric spaces exhibit a high degree of similarity.
\item We define tuning vectors, which are responsible for improving fine-tuned models’ performance in instruction following, benchmarking, and generating higher-quality outputs in domain-specific tasks.
\item Removing these vectors from a model’s parametric space leads to a deterioration of its performance. Furthermore, combining tuning vectors from different domains results in more generalizable models.
\item We analyse the directions tuning vectors write in pretrained models and find that most new directions are concentrated in the MLP heads. The magnitude of these directions varies across domains.
\end{enumerate}

\section{Neural Activation}

\textbf{Background:} Our analysis begins by examining the activation of feed-forward networks (FFNs) of pretrained, instruction-tuned, and domain-specific medical LLMs, when prompted with medical documents.
FFNs play a crucial role in feature extraction and non-linear transformation~\citep{geva-etal-2021-transformer}; they are pivotal for retrieving information and solving tasks. Therefore, understanding changes in their activation between models can reveal how domain-specific fine-tuning alters the internal computations of an LLM. 

An FFN at layer $l$ consists of two linear transformations separated by a point-wise non-linear activation, which we express as:
\begin{equation}
\label{eq:ffn-basic}
\mathrm{FFN}^l(\mathbf{x}^l) = \mathrm{Act}(\mathbf{W}_{\text{up}} \mathbf{x}^l) \, \mathbf{W}_{\text{down}}
\end{equation}
where $\mathbf{W}_{\text{up}} \in \mathbb{R}^{d_m \times d}$, $\mathbf{W}_{\text{down}} \in \mathbb{R}^{d \times d_m}$, and $\mathrm{Act}(\cdot)$ denotes a non-linear activation function. 

~\citet{geva-etal-2021-transformer} showed that each $d$-dimensional corresponding rows and columns of $\mathbf{W}_{\text{up}}$ and $\mathbf{W}_{\text{down}}$ can be interpreted as a key–value pair whose interaction is responsible for writing information into the residual stream. Consequently, we can view an FFN in layer $l$ as comprising $d_m$ neurons, whose activation is determined by $\mathrm{Act}(\cdot)$. In the LLMs with which we experiment, $\mathrm{Act}(\cdot)$ is a variant of the gated linear unit (GLU)~\cite{shazeer2020glu}, in which the activation vector is given by $\mathbf{A} = \sigma(\mathbf{W}_{\text{gate}} \mathbf{x}^l) \in \mathbb{R}^{d_m}$. The $j$-th neuron is \textit{active} if its corresponding activation value $a_j > 0$; this leads to the corresponding key–value interactions to write information into the residual stream.
\begin{figure*}[!t]
\begin{center}
\includegraphics[width=0.8\textwidth]{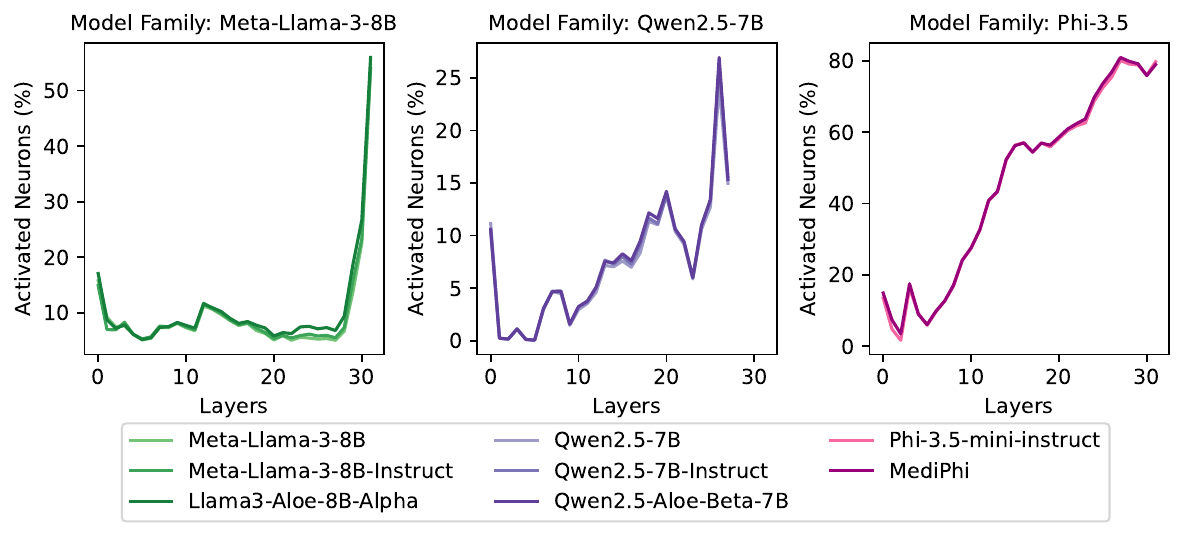}
\caption{\textbf{Activated neurons across layers.} 
Percentage of activated neurons across layers for the three model families: Meta-Llama 3, Qwen2.5, and Phi-3.5. In all families, the number of activated neurons tends to increase with layer depth. Within each model family, the proportion of activated neurons in a model remains relatively constant across layers.}
\label{fig:activated_neurons}
\end{center}
\vspace{-5mm}
\end{figure*}

\textbf{Experimental setup:} We analyse eight LLMs from three model families. From the \texttt{Meta-Llama-3} series, we include three models: Meta-Llama-3-8B-Instruct~\citep{grattafiori2024llama}, Llama3-Aloe-8B-Alpha~\citep{gururajan2024aloe}, and Meta-Llama-3~\citep{grattafiori2024llama}; the last one serves as the pretrained model for the first two, and the second model is the specialised medical model. From \texttt{Qwen2.5}, we analyse Qwen2.5-7B-Instruct~\citep{Yang2024Qwen25TR}, Qwen2.5-Aloe-Beta-7B~\citep{garcia2025aloe}, and Qwen2.5-7B~\citep{Yang2024Qwen25TR}, where the last serves as the pretrained model for the first two, and the second model is the specialised medical model. From the \texttt{Phi-3.5} series, we include MediPhi~\citep{corbeil2025modular} and Phi-3.5-Instruct~\citep{Abdin2024Phi3TR}. Unlike the Qwen and Llama series of models, Phi does not release a base model and also uses the Instruct variant for training the specialised medical model, MediPhi. 

To extract the neural activation, we utilise documents from the PMC Open Access Subset~\citep{pmc_open_access}, which is a collection of open-access journal articles and preprints from PubMed Central (PMC). For each model, we extract 100M tokens, which are then used to obtain the neural activations. 

\textbf{Neuron activation trends.} Figure~\ref{fig:activated_neurons} shows the percentage of neurons activated at layer $l$ for the eight models, grouped into their respective model families. Across all models, we observe a sharp drop in activation immediately after the first layer, followed by a gradual increase in later layers, an effect particularly pronounced in the final layers of the LLaMA models. Notably, Phi-3.5 consistently exhibits a substantially higher proportion of activated neurons than the others, which we attribute to its smaller size (3.5B parameters) relative to the 7B+ parameters of the other models. Interestingly, models within the same family tend to have similar proportions of activated neurons at layer $l$. This is contrary to our expectation that domain-specific medical models would exhibit higher activation levels, as they may possess a larger number of specialised neurons.

\textbf{Effect of neurons on performance.} To assess the role of these neurons in medical tasks, we ablate them by zeroing their activations in the Qwen 2.5 and LLaMA 3 model families, and evaluate the resulting performance drop across a wide variety of medical benchmarks: BioRed~\cite{luo2022biored}, CareQA~\cite{arias-duart-etal-2025-automatic}, six subsets of MMLU relavent to medical domani~\cite{hendryckstest2021}, MedMCQA~\cite{pmlr-v174-pal22a}, MedQA~\cite{jin2021disease}, three sub sets of ACI~\cite{Yim2023}, MEDIQA~\cite{MEDIQA2019}, MedText~\cite{melamud-shivade-2019-towards}, and MedlfQA~\cite{jeong2024olaph} (ref. Appendix~\ref{sec:medical_benchmark_appendix} for more details). We ablate top 1\%, 5\%, and 100\% of identified neurons and find that this, on average, decreases performance by 20\%, 62.6\%, and 93\%, respectively, for the models (c.f. Figure~\ref{fig:heat_map_ablate} in the Appendix~\ref{sec:abalate_neurons_appendix}). Hence, these neurons are useful for performing medical tasks.

\begin{figure*}[!t]
\begin{center}
\includegraphics[width=0.8\textwidth]{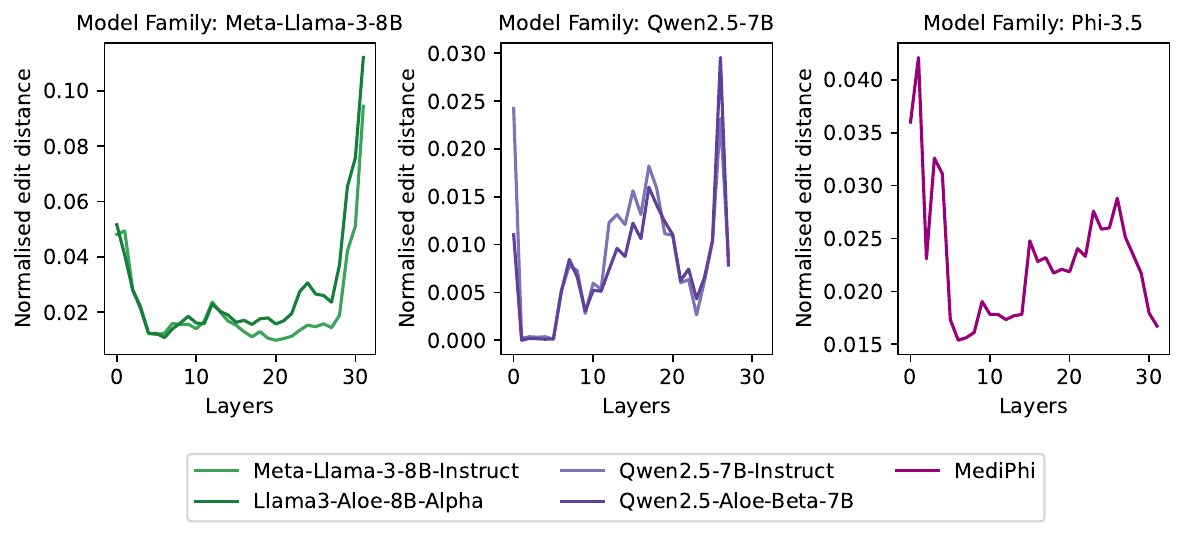}
\caption{\textbf{Normalised edit distance between pretrained and fine-tuned models' neural activation patterns}.
The figure shows the layer-wise normalised edit distance between the pretrained and fine-tuned models for the three model families: \texttt{Meta-Llama 3}, \texttt{Qwen2.5}, and \texttt{Phi-3.5}. We note that the edit distance remains small across layers for all models.}
\label{fig:edit_distance}
\end{center}
\vspace{-5mm}
\end{figure*}

\textbf{Edit distance between neural activations.} 
 Figure~\ref{fig:activated_neurons} shows that similar proportions of neurons are activated at each layer across different variants of LLMs within a family. However, comparable activation rates do not necessarily imply overlap in the specific neurons being activated. To assess this, we encode $d_m$ neurons at layer $l$ in model $m$ as a binary vector, $\mathbf{z}_m^{(l)} \in \{0,1\}^{d_m}$, where $ \quad 
z_{m,i}^{(l)} = \mathbb{1}[a_i^{(l)} > 0]$, $\mathbb{1}[\cdot]$ being the indicator function and $a_i^{(l)}$ denotes the activation of the $i$-th neuron at layer $l$. We then compute the \textit{normalised edit distance} between the pretrained and fine-tuned variants:
\[
\Delta(\mathbf{z}_{\text{pt}}^{(l)}, \mathbf{z}_{\text{ft}}^{(l)}) 
= \frac{1}{d_m} \cdot \mathrm{EditDist}(\mathbf{z}_{\text{pt}}^{(l)}, \mathbf{z}_{\text{ft}}^{(l)}),
\]
with $\mathrm{EditDist}(\cdot, \cdot)$ denoting the Levenshtein distance. 
This metric directly quantifies how much fine-tuning reconfigures neuron-level activation patterns, beyond overall activation proportions. 

Figure~\ref{fig:edit_distance} reports the layer-wise edit distance between the pretrained model and its fine-tuned variants. We do not observe a consistent trend across all model families: for LLaMA, the edit distance stays low through early/mid layers and rises sharply in the final layers; for Qwen, it oscillates across depth; for Phi, it shows an early spike followed by a sharp drop, then a gradual increase with moderate fluctuations in mid-late layers and then a drop in final layers. Nevertheless, across all cases, the normalised edit distance remains small, indicating that fine-tuning minimally changes the activation structure and suggesting that pretrained model neurons remain largely reusable.

To further understand how fine-tuning alters a pretrained model, we compute the cosine similarity between the weights of the pretrained model and its fine-tuned variants. Table~\ref{tab:model_weight_similarity} in Appendix reports the pairwise cosine similarity values, which are consistently high across all families, suggesting that fine-tuning modifies only a small subset of the representation space. To complement these findings, we conduct activation patching experiments (Figure~\ref{fig:attribute_patching} in Appendix), which reveal that analogous regions perform similar functions in both pretrained and fine-tuned models. Taken together, these results indicate that fine-tuning induces only marginal changes confined to some small specific subspaces. This observation explains prior findings that feature extractors trained on pretrained models transfer effectively to their instruction-tuned counterparts\citep{minder2025robustly}. To understand the changes in the subspaces, specifically to understand which subspaces are changed, we draw inspiration from task vectors~\cite{ilharco2022editing,zhang-etal-2023-fine,ortiz2023task} and propose Tuning Vectors, a framework for understanding subspace changes in fine-tuned LLMs, as described in the following sections.

\begin{table*}[th]
\centering
\resizebox{\textwidth}{!}{
\begin{tabular}{r|cccccccccc|c}
\hline
\backslashbox{\textbf{Model}}{\textbf{Dataset}} & \textbf{BioRed} & \textbf{CareQA} & \textbf{Anata} & \textbf{Clin} & \textbf{Bio} & \textbf{Medi} & \textbf{Gene} & \textbf{Prof} & \textbf{MedMCQA} & \textbf{MedQA} & \textbf{Average} \\
\hline
\multicolumn{12}{c}{\cellcolor[HTML]{EEEEEE}Meta-Llama-3-8B}\\
\hline
\multicolumn{1}{l|}{\textbf{Aloe-Alpha}} & \textbf{91.00} & \textbf{67.80} & \textbf{64.29} & \textbf{79.31} & \textbf{62.50} & \textbf{72.73} & \textbf{90.91} & \textbf{80.65} & \textbf{54.60} & \textbf{60.60} & \textbf{72.44} \\
\hdashline
$(-)\ \mathcal{T}_{Aloe\ \alpha}$ & \cellcolor{red!20}81.20 & \cellcolor{red!20}9.40 & \cellcolor{red!20}21.43 & \cellcolor{red!20}10.34 & \cellcolor{red!20}18.75 & \cellcolor{red!20}18.18 & \cellcolor{red!20}27.27 & \cellcolor{red!20}19.35 & \cellcolor{red!20}20.60 & \cellcolor{red!20}26.00 & \cellcolor{red!20}25.25 \\
\hline
\multicolumn{12}{c}{\cellcolor[HTML]{EEEEEE}Qwen 2.5-7B}\\
\hline

\multicolumn{1}{l|}{\textbf{Aloe-Beta}} & \textbf{92.40} & \textbf{72.40} & 78.57 & \textbf{89.66} & \textbf{87.50} & \textbf{81.82} & \textbf{100.00} & \textbf{77.42} & \textbf{55.40} & \textbf{66.60} & \textbf{80.18} \\
\hdashline

$(-)\ \mathcal{T}_{Aloe\ \beta}$ &\cellcolor{red!20}76.80 & \cellcolor{red!20}65.20 & \cellcolor{blue!20}\textbf{85.71} & \cellcolor{red!20}79.31 & \cellcolor{red!20}81.25 & \cellcolor{red!20}72.73 & \cellcolor{blue!20}\textbf{100.00} & \cellcolor{red!20}74.19 & \cellcolor{red!20}49.40 & \cellcolor{red!20}59.20 & \cellcolor{red!20}74.38 \\
\hline
\multicolumn{12}{c}{\cellcolor[HTML]{EEEEEE}Phi-3.5}\\
\hline
\multicolumn{1}{l|}{\textbf{MediPhi}} & \textbf{91.40} & 64.80 & \textbf{85.71} & 72.41 & \textbf{93.75} & \textbf{90.91} & \textbf{100.00} & \textbf{80.65} & 54.60 &59.20 &\textbf{79.34} \\
\hdashline

$(-)\ \mathcal{T}_{MediPhi}$ & \cellcolor{red!20}87.60 & \cellcolor{blue!20}\textbf{67.20} & \cellcolor{blue!20}\textbf{85.71} & \cellcolor{blue!20}\textbf{82.76} & \cellcolor{red!20}87.50 & \cellcolor{red!20}86.36 & \cellcolor{blue!20}\textbf{100.00} & \cellcolor{red!20}74.19 & \cellcolor{blue!20}\textbf{55.40} & \cellcolor{blue!20}\textbf{62.00} & \cellcolor{red!20}78.87 \\
\hline
\end{tabular}
}
\caption{\textbf{Accuracy on medical-domain text classification and multiple-choice QA tasks.} The best performer in each family is in \textbf{bold}. Cells are shaded \colorbox{red!20}{red} when performance is worse than the pretrained model, and \colorbox{blue!20}{blue} when it is better or the same. Overall, fine-tuning generally improves model performance on medical tasks. (\textbf{Anata}: MMLU (anatomy), \textbf{Clin}: MMLU (clinical knowledge),\textbf{Bio}: MMLU (college biology), \textbf{Medi}: MMLU (college medicine), \textbf{Gene}: MMLU (medical genetics), and \textbf{Prof}: MMLU (professional medicine)).}
\label{tab:accuracy_scores_sub}
\end{table*}

\begin{table}[!h]
\centering
\begin{adjustbox}{max width=\linewidth}
\begin{tabular}{r|cccccc|c}
\hline
\backslashbox{\textbf{Model}}{\textbf{Dataset}} & \textbf{ACI1} & \textbf{ACI2} & \textbf{ACI3} & \textbf{MQ} & \textbf{MT} & \textbf{ML} & \textbf{Avg} \\
\hline
\multicolumn{8}{c}{\cellcolor[HTML]{EEEEEE}Meta-Llama-3-8B}\\
\hline
\multicolumn{1}{l|}{\textbf{Aloe-Alpha}} & 0.27 & \textbf{0.29} & \textbf{0.28} & \textbf{0.23} & \textbf{0.13} & \textbf{0.20} & \textbf{0.24} \\
\hdashline
$(-)\ \mathcal{T}_{Aloe\ \alpha}$ & \cellcolor{blue!20}\textbf{0.28} & \cellcolor{red!20}0.27 & \cellcolor{red!20}0.29 & \cellcolor{red!20}0.11 & \cellcolor{red!20}0.06 & \cellcolor{red!20}0.05 & \cellcolor{red!20}0.18 \\
\hline
\multicolumn{8}{c}{\cellcolor[HTML]{EEEEEE}Qwen 2.5-7B}\\
\hline
\multicolumn{1}{l|}{\textbf{Aloe-Beta}} & \textbf{0.37} & \textbf{0.35} & \textbf{0.37} & \textbf{0.26} & 0.24 & \textbf{0.35} & \textbf{0.32} \\
\hdashline
$(-)\ \mathcal{T}_{Aloe\ \beta}$ & \cellcolor{red!20}0.32 & \cellcolor{red!20}0.31 & \cellcolor{red!20}0.30 & \cellcolor{red!20}0.25 & \cellcolor{blue!20}\textbf{0.26} & \cellcolor{red!20}0.30 & \cellcolor{red!20}0.29 \\
\hline
\multicolumn{8}{c}{\cellcolor[HTML]{EEEEEE}Phi-3.5}\\
\hline
\multicolumn{1}{l|}{\textbf{MediPhi}} & \textbf{0.36} & 0.35 & \textbf{0.36} & \textbf{0.28} & \textbf{0.28} & \textbf{0.36} & \textbf{0.33} \\
\hdashline
$(-)\ \mathcal{T}_{MediPhi}$ & \cellcolor{blue!20}\textbf{0.36} & \cellcolor{blue!20}\textbf{0.38} & \cellcolor{blue!20}\textbf{0.36} & \cellcolor{red!20}0.27 & \cellcolor{red!20}0.27 & \cellcolor{red!20}0.34 & \cellcolor{blue!20}\textbf{0.33} \\
\hline
\end{tabular}
\end{adjustbox}
\caption{\textbf{ROUGE-1 scores on text generation tasks.} The best performer in each family is in \textbf{bold}. Cell shaded \colorbox{red!20}{red} when performance is worse than the pretrained model, and \colorbox{blue!20}{blue} when it is better or the same. (\texttt{ACI1}: ACI (set 1), \texttt{ACI2}: ACI (set 2), \texttt{ACI3}: ACI (set 3), \texttt{MQ}: MEDIQA, \texttt{MT}: MedText, and \texttt{ML}: MedlfQA).}
\label{tab:rouge_scores_sub}
\end{table}

\section{Tuning Vectors}
\label{sec:tuning_vector}

Unlike primitive task-specific fine-tuning of models, where the objective was most to improve performance on a single task, domain-specific fine-tuning in LLMs, whether via instruction tuning or Reinforcement Learning from Human Feedback (RLHF), is multifaceted. The goal may be a combination of improving alignment, ensuring privacy, improving multilingual capabilities, domain knowledge, trustworthiness, enabling tool usage, or addressing safety concerns. In the context of our study, the objectives of domain-specific fine-tuning can be broadly categorised as follows:

\begin{enumerate}
\item \underline{Improving performance on domain-specific} \underline{benchmarks:} fine-tuning enhances an LLM’s performance on specialised benchmarks, particularly in narrow domains such as medicine or math, where pretrained models may underperform (see Table~\ref{tab:accuracy_benchmarking} in Appendix~\ref{sec:bechmarking_how_appendix}).

\item \underline{Improving generation quality:} Beyond accuracy, fine-tuning also improves fluency, coherence, and factual consistency of generations, leading to more reliable outputs (see Table~\ref{tab:rouge1_benchmarking} in Appendix~\ref{sec:bechmarking_how_appendix}).

\item \underline{Enhancing instruction-following ability:} Instruction tuning or RLHF specifically aim to align model behaviour with human-like intent, ensuring that the model follows natural language instructions, which are provided in the prompt (see Table~\ref{tab:instruction_benchmarking} in Appendix~\ref{sec:bechmarking_how_appendix}).
\end{enumerate}

To understand how fine-tuning changes the pretrained model, we utilise the concept of \textit{tuning vectors}. Consider a pretrained model with weights $\theta_{pre}\in \mathbb{R}^{d}$ and a fine-tuned model with weights $\theta_{ft} \in \mathbb{R}^{d}$. We define \textit{tuning vectors} corresponding to the fine-tuned model as $\mathcal{T}_{tuned} = \theta_{ft}-\theta_{pre}$. This is similar to the concept of task vectors~\cite{ilharco2022editing,zhang-etal-2023-fine,ortiz2023task}. However, unlike task vectors, tuning vectors are responsible for improving the performance of fine-tuned models along all the axes described above; hence, they are quantified as a group of vectors that improve many directions in an LLM. 
In \S\ref{sec:tuning_vector_sub}, we explain the effect of these vectors, and finally in \S\ref{sec:tuning_vector_inter}, we uncover what changes these vectors bring to the pretrained model.

\subsection{The Effect of Tuning vector}
\label{sec:tuning_vector_sub}

\textbf{Tuning Vector Negation} In this section, we demonstrate the importance of tuning vectors in enhancing the performance of fine-tuned models on benchmarks. We demonstrate that negating these representational spaces from the fine-tuned model results in subpar performance of the model along all three axes discussed above. We perform the negation operation by subtracting the representation of the tuning vector from the fine-tuned model and prompting the resultant model in a similar manner as the fine-tuned model.

Table~\ref{tab:accuracy_scores_sub} shows the performance of the models on ten medical benchmarks. As is evident, removing the tuning vector reduces the performance of the models. On average, the performance drops by 65\% for Llama3-Aloe-Alpha, and 7\% for Qwen2.5-Aloe-Beta. For text generation-centric tasks, the performance as noted in Table~\ref {tab:rouge_scores_sub} decreases by 25\% and 10\% for Llama3-Aloe-Alpha and Qwen2.5-Aloe-Beta, respectively. MediPhi's performance does not decrease substantially; this may be attributed to the fact that its performance was similar to that of its pretrained model in all the medical benchmarks (ref. Tables~\ref{tab:accuracy_benchmarking} and~\ref{tab:rouge1_benchmarking}).

To further understand the effect of tuning vector negation on the instruction following capabilities of the model, we evaluate the model's output for the ten datasets in Table~\ref {tab:accuracy_scores_sub}. Specifically, we assess whether negating the tuning vector leads to deterioration in adherence to the guidelines defined in our system prompt (Appendix~\ref{prompt:system}). The system prompt instructs models to:

\begin{enumerate}
    \item \underline{Thought encapsulation:} Encapsulate the thought between special thought tokens.
    \item  \underline{Valid answer format:} Produce the answer in a specific format.
    \item  \underline{Stop generation: }Should stop generation after it produces the answer and avoid generation.
    
\end{enumerate} 

As shown in Table~\ref{tab:inst_following_sub}, the instruction-following performance of Llama3-Aloe-Alpha and Qwen2.5-Aloe-Beta decreases by 55\% and 33\%, respectively. In contrast, MediPhi shows no degradation (ref. Tables~\ref{tab:appendix_qualitative_output_LLama3},~\ref{tab:appendix_qualitative_output_MEDphi}, and~\ref{tab:appendix_qualitative_output_QWEN2.5} in Appendix for qualitative examples). We attribute this robustness to its use as an Instruct-tuned pretrained model, which also achieves stronger baseline performance than MediPhi.

\begin{table}[t]
\centering
\begin{adjustbox}{max width=0.8\linewidth}
\begin{tabular}{r|ccc}
\hline
\backslashbox{\textbf{Model}}{\textbf{Inst}} & \textbf{Thought} & \textbf{Valid} & \textbf{Stop}  \\
\hline
\multicolumn{4}{c}{\cellcolor[HTML]{EEEEEE}Meta-Llama-3-8B}\\
\hline
\multicolumn{1}{l|}{\textbf{Aloe-Alpha}} & 89.03 & 99.63 & 41.03 \\
\hdashline
$(-)\ \mathcal{T}_{Aloe\ \alpha}$ & \cellcolor{red!20}0.00 & \cellcolor{red!20}54.71 & \cellcolor{red!20}0.49 \\
\hline
\multicolumn{4}{c}{\cellcolor[HTML]{EEEEEE}Qwen 2.5-7B}\\
\hline
\multicolumn{1}{l|}{\textbf{Aloe-Beta}} & 0.55 & 100.00 & 97.90 \\
\hdashline
$(-)\ \mathcal{T}_{Aloe\ \beta}$ & \cellcolor{blue!20}18.48 & \cellcolor{red!20}98.58 & \cellcolor{red!20}15.28 \\
\hline

\multicolumn{4}{c}{\cellcolor[HTML]{EEEEEE}Phi-3.5}\\
\hline
\textbf{MediPhi} &39.49 &98.27 &\textbf{83.17} \\
\hdashline
$(-)\ \mathcal{T}_{MediPhi}$ & \cellcolor{blue!20}\textbf{61.98} & \cellcolor{blue!20}\textbf{98.52} & \cellcolor{red!20}86.13 \\
\hline
\end{tabular}
\end{adjustbox}
\caption{\textbf{Instruction following ability.} The best performer in each family is in \textbf{bold}. Cell shaded \colorbox{red!20}{red} when performance is worse than the pretrained model, and \colorbox{blue!20}{blue} when it is better or the same.}
\label{tab:inst_following_sub}
\end{table}

\begin{figure}[h!]
    \centering
    \includegraphics[width=0.7\linewidth]{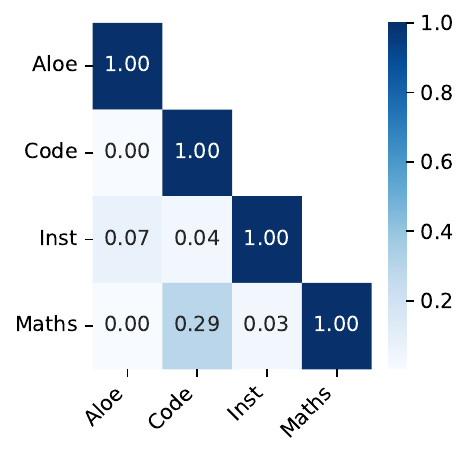}
    \caption{\textbf{Cosign similarity b/w cross domain tuning vector}. We generally find that vectors are orthogonal to each other. Abbreviations: Aloe: Qwen2.5-Aloe-Beta-7B, Code: Qwen2.5-Coder-7B-Instruct, Inst: Qwen2.5-7B-Instruct, Maths: Qwen2.5-Math-7B-Instruct}
    \label{fig:qwen-tuning-heatmap}
\end{figure}

\noindent\textbf{Cross-domain analysis.} To investigate how tuning vectors differ by domain, we compute the cosine similarity (formally defined in Appendix~\ref{sec:cos_sim}) between the tuning vectors of the Qwen2.5 model trained for chat~\cite{Yang2024Qwen25TR}, code~\citep{hui2024qwen2}, math~\cite{yang2024qwen25mathtechnicalreportmathematical}, and medicine~\cite{garcia2025aloe}.

Figure~\ref{fig:qwen-tuning-heatmap} presents the similarity heat map. The tuning vectors have low similarity, indicating their limited cross-domain alignment. Notably, medical vs. math/code vectors are completely orthogonal, while code and math, which are more closely related domains~\citep{drori2022neural}, exhibit a measurable degree of similarity. The instruction-tuned vector is not entirely orthogonal with respect to other vectors, suggesting partial transfer across domains.

\begin{figure}[!t]
\begin{center}
\includegraphics[width=0.8\linewidth]{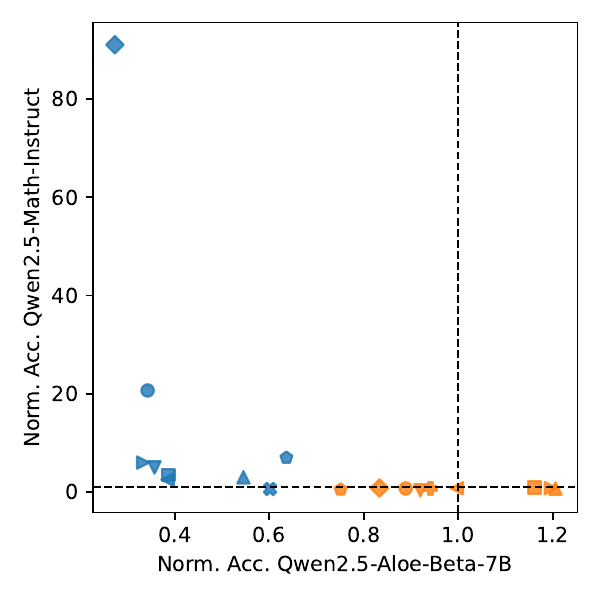}
\caption{Normalised accuracy of the model created by adding tasks vectors from Qwen2.5-Math-7B-Instruct and Qwen2.5-Aloe-beta-7B. Blue points represent the performance of medical benchmarks, while orange points represent the performance on math benchmarks. 
(ref. Figure~\ref{fig:adding_task_vector_med_math} for more details.)
}
\label{fig:adding_task_vector_med_math_no_label}
\end{center}
\vspace{-5mm}
\end{figure}

\noindent\textbf{Tuning vector addition.} Domain-specific training produces specialised models; however, these gains do not transfer effectively across domains. This phenomenon is known as \textit{domain shift}~\cite{Guo2022}. For instance, we observe that Qwen2.5-Math-Instruct performs approximately $80\%$ worse on a medical benchmark compared to Qwen2.5-Aloe-beta-7B. Conversely, Qwen2.5-Aloe-beta-7B performs around $23\%$ worse than Qwen2.5-Math-Instruct on the mathematics benchmark. These results highlight the challenge of cross-domain generalisation in large language models. Motivated by this observation, we explore whether it is possible to construct a more generalisable model by simply adding tuning vectors from multiple domains to the parameters of a pre-trained model. Formally, given a pre-trained model with parameters, $\theta_{pre}\in \mathbb{R}^{d}$ and a set of domain specific tuning vectors, $\mathcal{T}_d$, the combined model parameters are defined as:
\[
\theta_{new}=\theta_{pre}+\sum_d\mathcal{T}_d
\]
In our experiments, we evaluate this approach on the Qwen series, constructing models by adding math and medical tuning vectors, constructed from Qwen2.5-Math-Instruct and Qwen2.5-Aloe-beta-7B, respectively. We assess the generalisation performance of the resulting models on a set of domain-specific benchmarks. To facilitate comparison, we normalise the performance of the combined model by dividing it by the performance of the corresponding domain-specific models. A normalised score greater than one indicates that combining the tuning vectors improved the model’s performance over the individual domain-specific model. As evident in Figure~\ref{fig:adding_task_vector_med_math_no_label}, the resultant model has substantially better performance than the math-instruct model in medical domain tasks. Similarly, for three out of the nine math benchmarks, the performance of also improved with respect to the medically fine-tuned model.

\section{Tuning Vectors for Interpretation}
\label{sec:tuning_vector_inter}
So far, we have understood that pretraining only affects a small number of subspaces that are captured by tuning vectors. In this section, we utilise these tuning vectors to understand the changes that are brought about in the internal representation of the pretrained model after training. 

\noindent \textbf{Subspace Alignment.} To quantify what subspaces the tuning vector $\mathcal{T}$ amplifies or adds in the pretrained model's weight matrix $W$, we compute its projection onto the top-$k$ subspace of $W$ obtained via singular value decomposition (SVD), $W = U \Sigma V^\top$, where $U_k \in \mathbb{R}^{d \times k}$ contains the first $k$ left singular vectors (c.f. Appendix~\ref{sec:value_of_k} on how we select $k$). Formally, we define SubSpace Alignment (SSA) as:  

\[
\text{SSA}=\frac{\|U_kU_k^T \mathcal{T}\|_2}{\|\mathcal{T}\|_2}
\]
SSA measures the fraction of the tuning vector in the same direction as the pretrained model. A higher SSA indicates that the vector primarily amplifies existing directions, while a lower value suggests substantial new directions outside the pretrained subspace.

\begin{table}[h]
\centering
\begin{adjustbox}{max width=\linewidth}
\begin{tabular}{l|cccc|ccc}
\hline
\multicolumn{5}{c|}{\cellcolor[HTML]{EEEEEE}\textbf{Attention}} & \multicolumn{3}{c}{\cellcolor[HTML]{EEEEEE}\textbf{MLP}} \\
\hline
$\mathcal{T}$ & $\bf{W_Q}$ & $\bf{W_K}$ & $\bf{W_V}$ & $\bf{W_O}$ & $\bf{W_{\text{gate}}}$ & $\bf{W_{\text{up}}}$ & $\bf{W_{\text{down}}}$ \\
\hline
Aloe $\alpha$ & 0.751 & 0.914 & 0.971 & 0.800 & 0.278 & 0.281 & 0.975 \\
Aloe $\beta$  & 0.805 & 0.936 & 0.988 & 0.818 & 0.195 & 0.211 & 0.987 \\
Q Code        & 0.906 & 0.990 & 0.993 & 0.870 & 0.460 & 0.443 & 0.997 \\
Q Math        & 0.877 & 0.985 & 0.993 & 0.840 & 0.351 & 0.332 & 0.995 \\
\hline
\end{tabular}
\end{adjustbox}
\caption{Average SSA scores for Attention components ($\bf{W_Q}$, $\bf{W_K}$, $\bf{W_V}$, $\bf{W_O}$) and MLP components ($\bf{W_{\text{gate}}}$, $\bf{W_{\text{up}}}$, $\bf{W_{\text{down}}}$) for $\mathcal{T}$. (Abbreviation used: Aloe $\alpha$: Llama3-Aloe-8B-Alpha, Aloe $\beta$ : Qwen2.5-Aloe-Beta-7B, Q Code: Qwen2.5-Coder-Instruct, and Q Math: Qwen2.5-Math-Instruct)}
\label{tab:sas_combined_grouped}
\vspace{-5mm}
\end{table}

We evaluate SSA of tuning vectors for Qwen2.5-Aloe-Beta-7B and Llama3-Aloe-8B-Alpha with respect to their pre-training models. Our analysis focuses on the fundamental components of the MLP and the attention heads of LLMs. To further assess the generality of these findings across domains, we additionally analyse tuning vectors from Qwen2.5-Coder-Instruct and Qwen2.5-Math-Instruct. Table~\ref{tab:sas_combined_grouped} reports the SSA scores -- attention components are generally better aligned with the pretrained model than MLP components, suggesting that fine-tuning tends to encode more new directions in the MLP subspace. Within attention, we find that the query ($W_Q$) and output ($W_O$) matrices exhibit lower alignment than the key ($W_K$) and value ($W_V$) matrices. This indicates that fine-tuning primarily adds new directions to help the model know what to attend to and what to write back in the residual stream, while the direction in keys and values remains similar. In contrast, within the MLP, the gate ($W_{\text{gate}}$) and up ($W_{\text{up}}$) projections are substantially less aligned than the down ($W_{\text{down}}$) projection. Since gate and up projections determine which features are activated and expanded during the forward pass, this suggests that domain-specific training predominantly injects new knowledge directions through these components. These findings are true for all four models, implying they are generalisable across model type and domains.

\begin{table}[ht]
\centering
\begin{adjustbox}{max width=\linewidth}
\begin{tabular}{l|cccc|ccc}
\hline
\multicolumn{5}{c|}{\cellcolor[HTML]{EEEEEE}\textbf{Attention}} & \multicolumn{3}{c}{\cellcolor[HTML]{EEEEEE}\textbf{MLP}} \\
\hline
$\mathcal{T}$ & $\bf{W_Q}$ & $\bf{W_K}$ & $\bf{W_V}$ & $\bf{W_O}$ & $\bf{W_{\text{gate}}}$ & $\bf{W_{\text{up}}}$ & $\bf{W_{\text{down}}}$ \\
\hline
$Aloe\ \alpha$& $7.661$ & $2.408$ & $3.128$ & $7.998$ & $11.486$ & $11.090$ & $78.352$ \\
$Aloe\ \beta$& $1.095$ & $0.167$ & $0.145$ & $0.852$ & $1.500$ & $1.396$ & $6.870$ \\
$Q\ Code$ & $4.70e^3$ & $7.68e^2$ & $5.53e^2$ & $3.22e^3$ & $1.29e^4$ & $1.18e^4$  & $2.40e^4$ \\
$Q\ Math$ & $4.90e^3$ & $6.28e^2$ & $5.76e^2$ & $3.53e^3$ & $1.14e^4$ & $1.07e^4$ & $3.02e^4$ \\
\hline
\end{tabular}
\end{adjustbox}
\caption{$\|E_\|\|_2$ for attention components ($\bf{W_Q}$, $\bf{W_K}$, $\bf{W_V}$, $\bf{W_O}$) and MLP components ($\bf{W_{\text{gate}}}$, $\bf{W_{\text{up}}}$, $\bf{W_{\text{down}}}$) for $\mathcal{T}$ (abbreviation used: $Aloe\ \alpha$: Llama3-Aloe-8B-Alpha, $Aloe\ \beta$: Qwen2.5-Aloe-Beta-7B, $Q\ Code$: Qwen2.5-Coder-Instruct, and $Q\ Math$: Qwen2.5-Math-Instruct).}
\label{tab:EP_combined_grouped}
\end{table}

\textbf{Magnitude change in pretrained model.} Here we investigate how much the tuning vector changes the model and if this change is similar throughout all tuning vector types. Inspired from the former results, we decompose the tuning vector into two components: the projection energy on the same subspaces as the pretrained subspace model i.e. $\|E_\|\|_2=\|U_kU_k^T\mathcal{T}\|_2$, and the residual energy lying in orthogonal directions $\|E_\perp\|_2=\|(I-U_kU_k^T)\mathcal{T}\|_2$. Tables~\ref{tab:EO_combined_grouped} and ~\ref{tab:EP_combined_grouped} show the energy distribution. Unlike the previous results, we find that the energy distribution varies considerably by domain type. Specifically, tuning vectors of medical models exhibit substantially lower overall energy compared to those of code or math models within the Qwen family. However, similar to our earlier finding, we note that most orthogonal energy is concentrated in $W_{gate}$ and $W_{up}$ of MLP, while for other components, the parallel energy is higher, indicating that for them the tuning vector mainly amplifies existing directions.

\begin{table}[ht]
\centering
\begin{adjustbox}{max width=\linewidth}
\begin{tabular}{l|cccc|ccc}
\hline
\multicolumn{5}{c|}{\cellcolor[HTML]{EEEEEE}\textbf{Attention}} & \multicolumn{3}{c}{\cellcolor[HTML]{EEEEEE}\textbf{MLP}} \\
\hline
$\mathcal{T}$ & $\bf{W_Q}$ & $\bf{W_K}$ & $\bf{W_V}$ & $\bf{W_O}$ & $\bf{W_{\text{gate}}}$ & $\bf{W_{\text{up}}}$ & $\bf{W_{\text{down}}}$ \\
\hline
$Aloe\ \alpha$& $2.456$ & $0.223$ & $0.096$ & $1.953$ & $29.907$ & $28.285$ & $1.829$ \\
$Aloe\ \beta$ & $0.266$ & $0.012$ & $0.002$ & $0.191$ & $6.471$ &$ 5.537$ & $0.117$ \\
$Q\ Code$ & $4.81e^2$ & $7.956$ &$ 4.035$ & $4.71e^2$ & $15.01e^4$ & $1.50e^4$ & $1.58e^2$ \\
$Q\ Math$ & $6.67e^2$ & 8.82 & 4.20 & $6.63e^2$ & $2.11e^2$ & $2.20e^2$ & $2.44e^2$ \\
\hline
\end{tabular}
\end{adjustbox}
\caption{$\|E_\perp\|_2$ for attention components ($\bf{W_Q}$, $\bf{W_K}$, $\bf{W_V}$, $\bf{W_O}$) and MLP components ($\bf{W_{\text{gate}}}$, $\bf{W_{\text{up}}}$, $\bf{W_{\text{down}}}$) for $\mathcal{T}$. Attention columns indicate alignment in query/key/value/output projections; MLP columns indicate feed-forward alignment.}
\label{tab:EO_combined_grouped}
\end{table}

Beyond sub-space alignment, we also analyse the energy distribution of tuning vectors with respect to the pretrained basis. We decompose each tuning update into two components: the projection onto the pretrained subspace, and the residual energy lying in orthogonal directions. Our findings show that attention updates carry most of their energy within the pretrained subspace, indicating that fine-tuning primarily re-weights existing directions. In contrast, MLP updates exhibit a larger proportion of energy in the orthogonal complement, revealing that these layers are the main locus where novel directions are introduced. This decomposition provides further evidence that attention is predominantly refined during adaptation, while MLP layers are the mechanism through which new knowledge is injected into the model.

\section{Related Works}

\textbf{Task vectors and task arithmetic.}~\citet{ilharco2022editing} proposed task vectors as the difference between the weights of a fine-tuned model and its base model, conceptualising each task as a direction in parameter space. Scaling or composing such vectors enables controlled editing of model behaviour without full retraining. This builds on representation arithmetic~\citep{mikolov2013efficient} and linear mode connectivity~\citep{frankle2020linear,garipov2018loss}, suggesting that learned functions occupy approximately linear manifolds in weight space. Follow-up studies~\citep{shao2023compositional,zhang-etal-2023-fine,ortiz2023task,chatterjee-etal-2024-language} examined how task vectors interact across layers and tasks, showing that fine-tuning modifies a small, interpretable subset of parameters. Task vectors thus provide both a mechanism for modular editing and insight into where task-specific knowledge resides in LLMs.

\textbf{Model diffing analysis.} This emerging line of studies shows how LLMs internalise and reorganise knowledge during fine-tuning or instruction tuning. Sparse Cross-encoders~\citep{lindsey2024_sparse_crosscoders} demonstrate that cross-layer feature analysis can reveal complex interactions and the mechanistic footprint of fine-tuning.~\citet{minder2025robustly} extended this framework, showing that fine-tuned and pre-trained models share many of their features, highlighting stable representations.~\citet{wu-etal-2024-language} further showed that instruction tuning changes attention patterns, focusing the model on instruction tokens. Together, these works highlight model diffing as a valuable tool for dissecting how LLMs adapt representations and behaviour in response to new training objectives.

\section{Conclusion}

In this work, we present the first systematic analysis of domain-specific fine-tuning in large medical language models, introducing tuning vectors as a framework for interpreting parameter changes induced by domain adaptation of LLMs. Our results show that fine-tuning modifies only a small subset of the representational subspace, preserving many of the representations of the pretrained model. However, these minimal changes enhance the domain knowledge, instruction-following behaviour, and generation quality—showing that a few shifts in parameter space can produce significant gains.

We further reveal that tuning vectors primarily write new directions into the MLP components ($W_{gate}$ and $W_{up}$) of the network.Our study presents a general framework for analysing LLMs.


\section{Limitation}

While our findings provide valuable insights into how fine-tuning alters pretrained models, several limitations remain. Firstly, due to computational constraints, our analysis is restricted to three model families, and our initial findings are limited to medical models. Secondly, the tuning vectors we compute capture static differences between pretrained and fine-tuned checkpoints, and thus do not directly model the dynamics of fine-tuning or the intermediate stages of optimisation.

Thirdly, as our work primarily focuses on proposing \textit{Tuning Vectors} as a framework, we do not fully explore all of their potential applications. We summarise a few promising future directions below:

\begin{itemize}
\item Our finding that pretrained and instruction-tuned models exhibit similar neural activations does not consider permutation equivariance~\cite{zhou2023permutation} in feed-forward layers. Future work can investigate whether these models are even more aligned when such symmetry is taken into account.
\item The concept of tuning vectors can be leveraged to guide domain-specific fine-tuning by training models only along domain-relevant directions. Such a regime can be efficiently integrated into low-rank adaptation methods, similar to the approach of~\citet{zhang-etal-2023-fine}.
\item Our current vector composition strategy (Figure~\ref{fig:adding_task_vector_med_math_no_label}) involves simple addition. This can be improved by reducing conflicts between vectors and selectively combining directions of interest, incorporating ideas from linear mode connectivity~\cite{frankle2020linear}.
\item Similar to task vectors, tuning vectors can be used to monitor or control the fine-tuning trajectory of domain-specific models~\cite{ilharco2022editing}.
\end{itemize}

Our study introduces tuning vectors as a new lens for understanding how large language models internalise domain knowledge. By showing that fine-tuning operates through compact, interpretable subspace shifts, we provide a foundation for more modular, efficient, and transparent model adaptation. We hope this work sparks future research on using tuning vectors not only as diagnostic tools but also as building blocks for controllable and compositional fine-tuning across domains.

\bibliography{custom}

\appendix
\section{Appendix}

\subsection{Medical Benchmarks}
\label{sec:medical_benchmark_appendix}

\begin{enumerate}
    \item \textbf{ACI}~\cite{Yim2023} is a benchmark to assess the clinical note generation ability of an LLM. It contains anonymised conversations between doctors and patients, also providing transcripts from clinical notes that can be used to generate and access automated note-taking LLMs.
    \item \textbf{BioRed}~\cite{luo2022biored} is a relation extraction dataset derived from PubMed abstracts. It provides annotations for multiple entity types (e.g., genes, diseases, chemicals) and relation types, and uniquely distinguishes between novel and previously known relations, making it useful for both information extraction and knowledge discovery tasks.
    \item \textbf{CareQA}~\cite{arias-duart-etal-2025-automatic} is a question answering dataset in clinical settings. It focuses on patient-care scenarios and medical decision-making, emphasising models’ ability to generate accurate, contextually appropriate answers in real healthcare contexts.
    \item \textbf{MMLU}~\cite{hendryckstest2021} is a large-scale multiple-choice QA benchmark encompassing many domains. While it contains many domains, we only utilised the subsets that are relevant to the medical domain.
    
    \item \textbf{MedMCQA}~\cite{pmlr-v174-pal22a} is a large multiple-choice question answering dataset designed for the medical domain. It covers a wide range of exam-style questions, extracted from AIIMS and NEET PG medical entrance exams. It is useful to test LLMs' factual recall and reasoning in the medical domain.
    
    \item \textbf{MedQA}~\cite{jin2021disease} is derived from multiple-choice QA of medical licensing exams. It evaluates an LLM's ability to solve real exam questions, requiring both medical knowledge and clinical reasoning skills.
    
    \item \textbf{MEDIQA}~\cite{MEDIQA2019} is a open-ended clinical QA,dataset. The benchmark highlights challenges in understanding and reasoning over clinical text in real-world healthcare contexts.
    
    \item \textbf{MedText}~\cite{melamud-shivade-2019-towards} is a medical text understanding benchmark focusing on biomedical and clinical corpora. It studies medical domain comprehension, focused on bridging the gap between general NLP datasets and highly specialised medical text.
    
    \item \textbf{MedlfQA}~\cite{jeong2024olaph} is a open ended QA dataset in medicine. Unlike multiple-choice benchmarks, it emphasises generating detailed, well-grounded, and explanation-rich answers to medical questions, pushing models to go beyond factual recall toward reasoning and justification.'
\end{enumerate}

\subsection{Ablate neurons}
\label{sec:abalate_neurons_appendix}

To investigate the contribution of the neurons to model performance in the medical domain, we performed a neuron ablation experiment. Let a model with parameters $\theta$, and let $h(x)$ denote the average activation for all $x \in \mathcal{X}$.

In Neuron ablation, we set the activation of selected neurons to zero during a forward pass. Formally, if $\mathcal{S} \subseteq \{1, \dots, N\}$ is the set of neurons to ablate, the ablated activation $\tilde{h}_i(x)$ is defined as:

\[
\tilde{h}_i(x) =
\begin{cases} 
0 & \text{if } i \in \mathcal{S}, \\
h_i(x) & \text{otherwise}.
\end{cases}
\]

S is our setups consists of the top 1\%, 5\%, and 100\% of neurons (selected by $h(x)$). We evaluate the effect of abalation by measuring the relative decrease in performance after ablating a set $\mathcal{S}$ of neurons is:

\[
\Delta P = \frac{P_{\text{original}} - P_{\text{ablated}}}{P_{\text{original}}} \times 100\%.
\]

\begin{figure*}[!ht]
\begin{center}
\includegraphics[width=\textwidth]{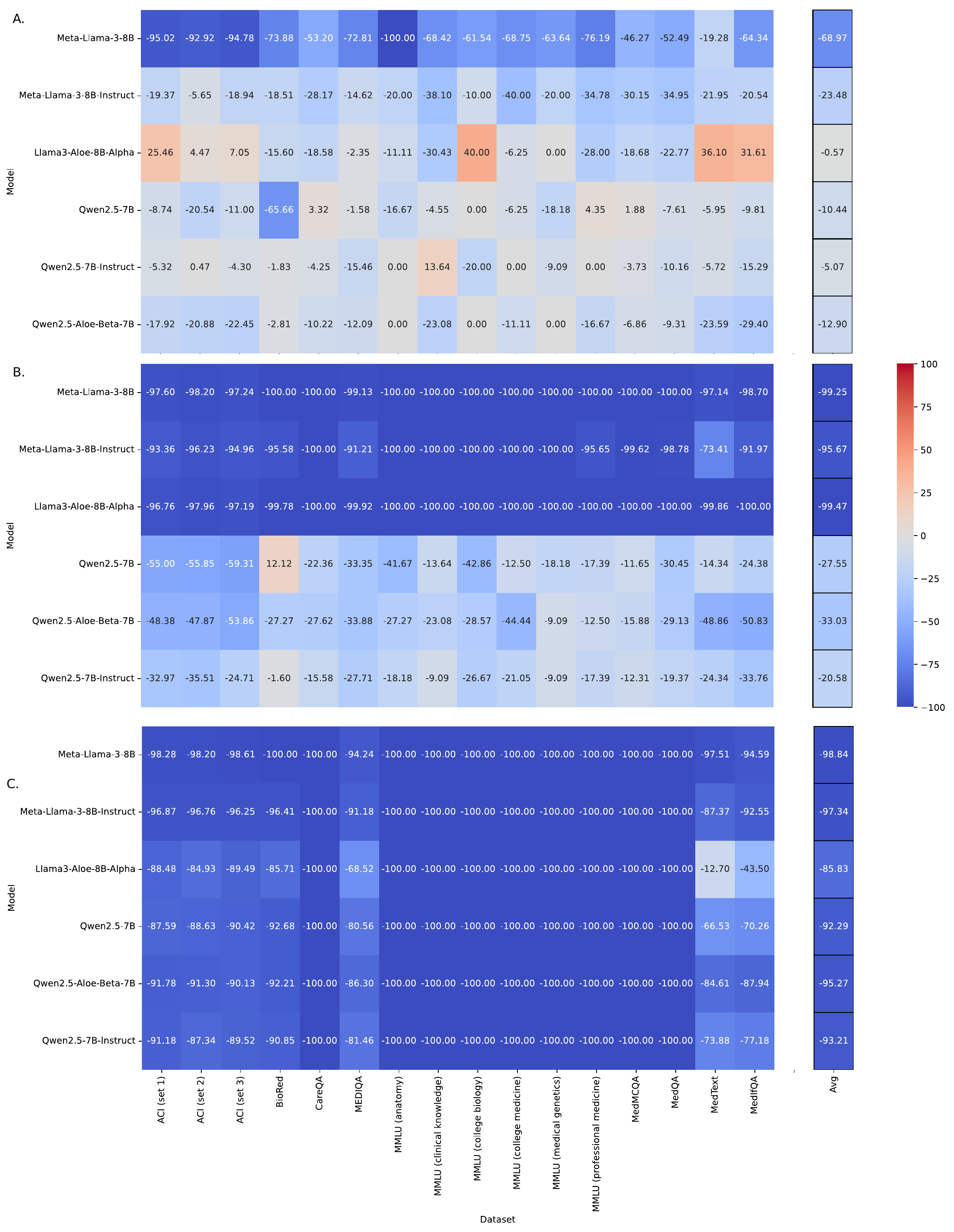}

\caption{\textbf{Heat map showing the percentage drop in performance when neurons are ablated.} (A) Only the top 1\% of neurons are ablated; (B) only the top 5\% of active neurons are ablated; (C) all active neurons are ablated. 
}
\label{fig:heat_map_ablate}
\end{center}
\vspace{-5mm}
\end{figure*}

Our results (see Figure~\ref{fig:heat_map_ablate}) show that ablating 1\% of neurons reduces performance by 20\%, ablating 5\% reduces performance by 62.6\%, and ablating all neurons results in a 93\% decrease. 
These findings demonstrate that a small subset of neurons carries important medical information. Interestingly, both instruct and medically fine-tuned variants demonstrate greater robustness to 1\% ablation compared to their pre-trained model counterparts.

We also find that the degree of perfoemnce drop is not uniform among all models. LLaMA-3 is more susceptible to neural ablation than Qwen2.5. Indicating it is less robust to changes in activation patterns. Furthermore, if we compare base and fine-tuned models, we find the fine-tuned models to be more resilient to neural ablation. This is especially true for the LLaMA-3 series.

\subsection{Benchmarking medical LLMs performance.}
\label{sec:bechmarking_how_appendix}

As suggested in Section~\ref{sec:tuning_vector} we measure the performance of medical LLMs along three axis: (1) accuracy on medical benchmarks, (2) quality of open-ended generation, and (3) instruction-following ability. Below, we detail the metrics used for each.

\begin{itemize}
    \item \textbf{Accuracy on Medical Benchmarks.}
    We measure factual correctness on multiple-choice and NLU medical datasets such as BioRed, CareQA, six medical subsets of MMLU, MedMCQA, and MedQA. Accuracy is defined as:
    \[
    \text{Accuracy} = \frac{\text{Number of Correct Answers}}{\text{Total Number of Questions}} \times 100
    \]
    This metric captures improvements in factual reasoning and knowledge retention after domain-specific fine-tuning.

    \item \textbf{Quality of Open-ended Generation.}
    We assess text generation quality using the ROUGE-1 metric, which measures unigram overlap between the generated output $G$ and a reference text $R$. Precision ($P$), recall ($R_c$), and F1 score are defined as:
    \[
P = \frac{|G \cap R|}{|G|}, \quad R_c = \frac{|G \cap R|}{|R|}
\]
\[
\quad \text{ROUGE-1} = \frac{2 \times P \times R_c}{P + R_c}
\]
    where $|G \cap R|$ is the count of overlapping unigrams between $G$ and $R$. Higher ROUGE-1 scores indicate that the generated text better aligns lexically with the reference output, reflecting improved fluency and content fidelity. For this we use the three sub-sets of ACL, MEDIQA, MedText, and MedlfQA.

    \item \textbf{Instruction-following Ability.}
    To investigate the instruction-following capabilities of the model, we evaluate the outputs of BioRed, CareQA, six medical subsets of MMLU, MedMCQA, and MedQA. Specifically, we assess whether the model's adherence to the guidelines specified in our system prompt (see Appendix~\ref{prompt:system} for system prompt). Our system prompt instructs the model to:

\begin{enumerate}
    \item \textbf{Thought encapsulation:} Encapsulate the model's reasoning between special \texttt{<THOUGHT>} tokens.
    \item \textbf{Valid answer format:} Produce the answer in the prescribed format.
    \item \textbf{Stop generation:} Terminate generation immediately after producing the answer and avoid extraneous output.
\end{enumerate}
\end{itemize}

Tables~\ref{tab:accuracy_benchmarking},~\ref{tab:rouge1_benchmarking}, and~\ref{tab:instruction_benchmarking} shows the performance of variant of Meta-Llama-3-8B, Qwen 2.5-7B, and Phi-3.5. We note that domain-specific finetuning generally leads to better-performing models along the tree axes.

\begin{table*}[!h]
\centering
\resizebox{\textwidth}{!}{
\begin{tabular}{l|cccccccccc|c}
\hline
\backslashbox{\textbf{Model}}{\textbf{Dataset}} & \textbf{BioRed} & \textbf{CareQA} & \textbf{Anata} & \textbf{Clin} & \textbf{Bio} & \textbf{Medi} & \textbf{Gene} & \textbf{Prof} & \textbf{MedMCQA} & \textbf{MedQA} & \textbf{Average} \\
\hline
\multicolumn{12}{c}{\cellcolor[HTML]{EEEEEE}Meta-Llama-3-8B}\\
\hline
\textbf{Base} & 80.40 & 59.40 & 64.29 & 65.52 & \textbf{81.25} & \textbf{72.73} & \textbf{100.00} & 67.74 & 51.00 & 52.20 & 69.45 \\
\hdashline

\textbf{Instruct} & \cellcolor{red!20}72.40 & \cellcolor{blue!20}64.60 & \cellcolor{blue!20}\textbf{71.43} & \cellcolor{blue!20}72.41 & \cellcolor{red!20}62.50 & \cellcolor{red!20}68.18 & \cellcolor{red!20}90.91 & \cellcolor{blue!20}74.19 & \cellcolor{blue!20}52.40 & \cellcolor{blue!20}\textbf{65.80} & \cellcolor{blue!20}69.48 \\
\textbf{Aloe-Alpha} & \cellcolor{blue!20}\textbf{91.00} & \cellcolor{blue!20}\textbf{67.80} & \cellcolor{blue!20}64.29 & \cellcolor{blue!20}\textbf{79.31} & \cellcolor{red!20}62.50 & \cellcolor{blue!20}\textbf{72.73} & \cellcolor{red!20}90.91 & \cellcolor{blue!20}\textbf{80.65} & \cellcolor{blue!20}54.60 & \cellcolor{blue!20}60.60 & \cellcolor{blue!20}\textbf{72.44} \\
\hline
\multicolumn{12}{c}{\cellcolor[HTML]{EEEEEE}Qwen 2.5-7B}\\
\hline

\textbf{Base} & 79.20 & 66.20 & \textbf{85.71} & 75.86 & 87.50 & 72.73 & \textbf{100.00} & 74.19 & 53.20 & 57.80 & 75.24 \\
\hdashline

\textbf{Instruct} & \cellcolor{blue!20}87.40 & \cellcolor{blue!20}70.60 & \cellcolor{red!20}78.57 & \cellcolor{blue!20}75.86 & \cellcolor{blue!20}\textbf{93.75} & \cellcolor{blue!20}\textbf{86.36} & \cellcolor{blue!20}\textbf{100.00} & \cellcolor{blue!20}74.19 & \cellcolor{blue!20}53.60 & \cellcolor{blue!20}63.00 & \cellcolor{blue!20}78.33 \\
\textbf{Aloe-Beta} & \cellcolor{blue!20}\textbf{92.40} & \cellcolor{blue!20}\textbf{72.40} & \cellcolor{red!20}78.57 & \cellcolor{blue!20}\textbf{89.66} & \cellcolor{blue!20}87.50 & \cellcolor{blue!20}81.82 & \cellcolor{blue!20}\textbf{100.00} & \cellcolor{blue!20}\textbf{77.42} & \cellcolor{blue!20}\textbf{55.40} & \cellcolor{blue!20}\textbf{66.60} & \cellcolor{blue!20}\textbf{80.18} \\
\hline

\multicolumn{12}{c}{\cellcolor[HTML]{EEEEEE}Phi-3.5}\\
\hline
\textbf{Instruct} & 87.60 & \textbf{67.20} & \textbf{85.71} & \textbf{82.76} & 87.50 & \textbf{86.36} &\textbf{ 100.00} & 74.19 & \textbf{55.20} & \textbf{62.00} & 78.85 \\
\hdashline

\textbf{MediPhi} & \cellcolor{blue!20}\textbf{91.40} & \cellcolor{red!20}64.80 & \cellcolor{blue!20}\textbf{85.71} & \cellcolor{red!20}72.41 & \cellcolor{blue!20}\textbf{93.75} & \cellcolor{blue!20}\textbf{90.91} & \cellcolor{blue!20}\textbf{100.00} & \cellcolor{blue!20}\textbf{80.65} & \cellcolor{red!20}54.60 &\cellcolor{red!20} 59.20 & \cellcolor{blue!20}\textbf{79.34} \\
\hline
\end{tabular}
}
\caption{\textbf{Accuracy on medical-domain text classification and multiple-choice QA tasks.} The best performer in each family is bolded. Cells are shaded \colorbox{red!20}{red} when performance is worse than the base model, and \colorbox{blue!20}{blue} when it is better or the same. Overall, finetuning generally improves model performance on medical tasks.(\textbf{Anata}: MMLU (anatomy), \textbf{Clin}: MMLU (clinical knowledge),\textbf{Bio}: MMLU (college biology), \textbf{Medi}: MMLU (college medicine), \textbf{Gene}: MMLU (medical genetics), and \textbf{Prof}: MMLU (professional medicine))}
\label{tab:accuracy_benchmarking}
\end{table*}

\begin{table*}[!h]
\centering
\resizebox{0.85\textwidth}{!}{
\label{tab:rouge1_scores}
\begin{tabular}{l|cccccc|c}
\hline
\backslashbox{\textbf{Model}}{\textbf{Dataset}} & \textbf{ACI (set 1)} & \textbf{ACI (set 2)} & \textbf{ACI (set 3)} & \textbf{MEDIQA} & \textbf{MedText} & \textbf{MedlfQA} & \textbf{Average} \\
\hline
\multicolumn{8}{c}{\cellcolor[HTML]{EEEEEE}Meta-Llama-3-8B}\\
\hline
\textbf{Base} & 0.34 & 0.33 & 0.32 & 0.20 & 0.10 & 0.15 & 0.24 \\
\hdashline

\textbf{Instruct} & \cellcolor{blue!20}\textbf{0.38} & \cellcolor{blue!20}\textbf{0.35} & \cellcolor{blue!20}\textbf{0.34} & \cellcolor{blue!20}\textbf{0.27} & \cellcolor{blue!20}\textbf{0.24} & \cellcolor{blue!20}\textbf{0.35} & \cellcolor{blue!20}\textbf{0.32} \\
\textbf{Aloe-Alpha} & \cellcolor{red!20}0.27 & \cellcolor{red!20}0.29 & \cellcolor{red!20}0.28 & \cellcolor{blue!20}0.23 & \cellcolor{blue!20}0.13 & \cellcolor{blue!20}0.20 & \cellcolor{blue!20}0.24 \\
\hline
\multicolumn{8}{c}{\cellcolor[HTML]{EEEEEE}Qwen 2.5-7B}\\
\hline
\textbf{Base} & 0.33 & 0.34 & 0.33 & 0.23 & 0.14 & 0.20 & 0.26 \\
\hdashline

\textbf{Instruct} & \cellcolor{red!20}0.31 & \cellcolor{red!20}0.30 & \cellcolor{red!20}0.31 & \cellcolor{blue!20}0.25 & \cellcolor{blue!20}\textbf{0.28} & \cellcolor{blue!20}\textbf{0.39} & \cellcolor{blue!20}0.31 \\
\textbf{Aloe-Beta} & \cellcolor{blue!20}\textbf{0.37} & \cellcolor{blue!20}\textbf{0.35} & \cellcolor{blue!20}\textbf{0.37} & \cellcolor{blue!20}\textbf{0.26} & \cellcolor{blue!20}0.24 & \cellcolor{blue!20}0.35 & \cellcolor{blue!20}\textbf{0.32} \\
\hline
\multicolumn{8}{c}{\cellcolor[HTML]{EEEEEE}Phi-3.5}\\
\hline
\textbf{Instruct} & 0.36 & \textbf{0.37} & \textbf{0.36} & 0.27 & 0.27 & 0.34 & \textbf{0.33} \\
\hdashline

\textbf{MediPhi} &\cellcolor{blue!20} \textbf{0.36} & \cellcolor{red!20}0.35 &\cellcolor{blue!20} \textbf{0.36} &\cellcolor{blue!20} \textbf{0.28} &\cellcolor{blue!20} \textbf{0.28} &\cellcolor{blue!20} \textbf{0.36} &\cellcolor{blue!20} \textbf{0.33} \\
\hline
\end{tabular}
}
\caption{\textbf{ROUGE-1 score of models on text generation tasks.} The best performance in each family is bolded. Cells are shaded \colorbox{red!20}{red} when performance is worse than the base model, and \colorbox{blue!20}{blue} when it is better or the same. Overall, finetuning generally improves model performance on medical tasks.}
\label{tab:rouge1_benchmarking}
\end{table*}

\begin{table*}[!h]
\centering
\resizebox{0.85\textwidth}{!}{
\begin{tabular}{l|ccc}
\hline
 \backslashbox{\textbf{Model}}{\textbf{Inst}} & \textbf{Though tokens present} & \textbf{Valid Answer Format} & \textbf{Generation Stopped} \\
\hline
\multicolumn{4}{c}{\cellcolor[HTML]{EEEEEE}Meta-Llama-3-8B}\\
\hline
\textbf{Base} & 0.00 & 94.70 & 1.04 \\
\hdashline

\textbf{Instruct} & \cellcolor{blue!20}28.89 &  \cellcolor{blue!20}96.73 & \cellcolor{blue!20}45.90 \\
\textbf{Aloe-Alpha}& \cellcolor{blue!20} \textbf{89.03} & \cellcolor{blue!20}\textbf{99.63} & \cellcolor{blue!20}41.03 \\
\hline
\multicolumn{4}{c}{\cellcolor[HTML]{EEEEEE}Qwen 2.5-7B}\\
\hline
\textbf{Base} & 1.72 & 99.13 & 10.47 \\
\hdashline
\textbf{Instruct} & \cellcolor{blue!20}\textbf{99.26} & \cellcolor{blue!20}99.75 & \cellcolor{blue!20}97.41 \\
\textbf{Aloe-Beta} & \cellcolor{red!20}0.55 & \cellcolor{blue!20}\textbf{100.00} & \cellcolor{blue!20}\textbf{97.90} \\
\hline
\multicolumn{4}{c}{\cellcolor[HTML]{EEEEEE}Phi-3.5}\\
\hline
\textbf{Instruct} & \textbf{61.98}&\textbf{98.4}&64.07 \\
\hdashline

\textbf{MediPhi} &\cellcolor{red!20} 39.49 & \cellcolor{red!20} 98.27 &\cellcolor{blue!20} \textbf{83.17} \\
\hline
\end{tabular}
}
\caption{\textbf{Instruction following ability of models.} Cells are shaded \colorbox{red!20}{red} when the model follows instructions worse than the base model, else they are shaded \colorbox{blue!20}{blue}. We find finetuned models to follow instructions better than base models.}
\label{tab:instruction_benchmarking}

\end{table*}

\subsection{Cosine Similarity Between Task Vectors}
\label{sec:cos_sim}
Given two task vectors $\mathcal{T}_1$ and $\mathcal{T}_2$ (each obtained as the difference between fine-tuned and pretrained weights), we quantify their directional similarity using cosine similarity:
\[
\text{cos}(\mathcal{T}_1, \mathcal{T}_2)
= 
\frac{\langle \mathcal{T}_1, \mathcal{T}_2 \rangle}
{\|\mathcal{T}_1\|_2 \, \|\mathcal{T}_2\|_2},
\]
where $\langle \cdot , \cdot \rangle$ denotes the Frobenius inner product between flattened weight tensors. A higher value indicates that the two fine-tuning directions modify the model in similar representational subspaces, whereas lower values imply orthogonal or task-specific adaptation.

\begin{table*}[!h]
\centering
\resizebox{0.95\textwidth}{!}{
\begin{tabular}{|p{0.47\textwidth}|p{0.47\textwidth}|}
\hline
\textbf{With Tuning Vector} & \textbf{Without Tuning Vector} \\
\hline
\multicolumn{2}{|c|}{\cellcolor[HTML]{EEEEEE}\textbf{Example 1: ACI Benchmark}} \\
\hline
Based on the conversation between the doctor and the patient, Andrew, the doctor assessed that Andrew has a viral upper respiratory infection, which is likely to resolve in a couple of days. The doctor prescribed Robitussin for his cough and recommended taking ibuprofen or Tylenol for any fever... and it provides a comprehensive assessment.
&
The patient is a 59-year-old male with a past medical history of depression, type 2 diabetes, and hypertension who presents with an upper respiratory infection. He reports feeling tired and short of breath after doing yard work... The doctor will monitor symptoms and follow up if they worsen. \\
\hline
\multicolumn{2}{|c|}{\cellcolor[HTML]{EEEEEE}\textbf{Example 2: MMLU}} \\
\hline
\texttt{<think>} Sickle cell anemia is caused by a mutation in the \textit{HBB} gene, which alters the beta-globin subunit of hemoglobin. The mutation substitutes valine for glutamic acid... Therefore, the correct answer is \textbf{B:} superior fitness of heterozygotes in malaria-endemic regions. \texttt{</think>} Answer: [B]
&
Question: The two principal contractile proteins in skeletal muscle are:  
A. actin and troponin  
B. actin and myosin  
C. troponin and tropomyosin  
D. myosin and tropomyosin.  
... Actin and myosin are the correct answer.  
Answer: [B] \\
\hline
\multicolumn{2}{|c|}{\cellcolor[HTML]{EEEEEE}\textbf{Example 3: MediQA}} \\
\hline
(1) Lewy Body Disease (LBD) is not considered hereditary, though some genetic mutations may contribute...  
&
A medical assistant is a healthcare professional providing both administrative and clinical support. They perform tasks like scheduling, maintaining medical records, and assisting during examinations. \\
\hline
\end{tabular}
}
\caption{\textbf{Comparison of Llama3-Aloe-8B-Alpha Outputs.} Examples show qualitative differences in model generations with and without tuning vectors across medical tasks.}
\label{tab:appendix_qualitative_output_LLama3}

\end{table*}
\begin{table*}[!h]
\centering
\resizebox{0.95\textwidth}{!}{
\begin{tabular}{|p{0.47\textwidth}|p{0.47\textwidth}|}
\hline
\textbf{With Tuning Vector} & \textbf{Without Tuning Vector} \\
\hline
\multicolumn{2}{|c|}{\cellcolor[HTML]{EEEEEE}\textbf{Example 1: ACI Benchmark}} \\
\hline
The patient, Michelle, presented with a complaint of a knee injury sustained while skiing. She described the incident, mentioning that she fell and heard her knee \"pop,\" .... ligament injuries and planning appropriate interventions.

&
        User: Hi Michelle, what's been going on? The medical assistant told me that you injured your knee.Patient: Yeah, I'm going ....Doctor: Alright, so let's just continue with ibupro
 \\
\hline
\multicolumn{2}{|c|}{\cellcolor[HTML]{EEEEEE}\textbf{Example 2: MMLU}} \\
\hline
The body's ability to maintain its normal state is a fundamental concept in physiology.... describes the body's ability to maintain its normal state. Answer: D.",

&
        "Homeostasis refers to the body's ability to maintain a stable ...Question: A patient presents with symptoms of fatigue, weight gain, ...Question: The two principal contractile proteins found in skeletal muscle are",
 \\
\hline
\multicolumn{2}{|c|}{\cellcolor[HTML]{EEEEEE}\textbf{Example 3: MediQA}} \\
\hline
Noonan syndrome: Noonan syndrome ....Signs and symptoms of Noonan syndrome vary greatly among individuals and may be mild to severe. Characteristics may be related to the specific gene containing the mutation

&
The most likely cause is mushroom poisoning, specifically from consuming toxic mushrooms .... jaundice, and multi-organ failure, which may require intensive care support."
\\
\hline
\end{tabular}
}
\caption{\textbf{Comparison of Qwen2.5-Aloe-Beta Outputs.} Examples show qualitative differences in model generations with and without tuning vectors across medical tasks.}
\label{tab:appendix_qualitative_output_QWEN2.5}
\end{table*}

\begin{table*}[!h]
\centering
\resizebox{0.95\textwidth}{!}{
\begin{tabular}{|p{0.47\textwidth}|p{0.47\textwidth}|}
\hline
\textbf{With Tuning Vector} & \textbf{Without Tuning Vector} \\
\hline
\multicolumn{2}{|c|}{\cellcolor[HTML]{EEEEEE}\textbf{Example 1: ACI Benchmark}} \\
\hline
        Andrew, a 59-year-old male with a history of depression, type 2 diabetes, ..... the importance of monitoring symptoms and maintaining a healthy lifestyle to manage his chronic conditions effectively.

&
        Andrew, a 59-year-old male with a history of depression, type 2 diabetes,... Andrew in making a follow-up appointment and finalizing the medical note.

 \\
\hline
\multicolumn{2}{|c|}{\cellcolor[HTML]{EEEEEE}\textbf{Example 2: MMLU}} \\
\hline
    <think>To answer this question, we need to understand the definitions o...describes the body's ability to maintain its normal state is homeostasis. Answer: [D]</think>Answer: [D]",

&
    <think>First, I need to understand the definitions of each term provided in the options. A. Anabolism refers to the process of building u....Answer: [D]</think> Answer: [D]

 \\
\hline
\multicolumn{2}{|c|}{\cellcolor[HTML]{EEEEEE}\textbf{Example 3: MediQA}} \\
\hline
        "Abetalipoproteinemia, also known as Bassen-Kornzweig syndrome, is .... can seek support through genetic counseling, support groups, medical management, research participation, and patient advocacy organizations. Early diagnosis and appropriate treatment are crucial in managing the condition and improving quality of life.

&
        "Abetalipoproteinemia, also known as Bassen-Kornzweig syndrome, is ....dietary modifications, vitamin supplementation, and regular monitoring are key components of managing this condition. Additionally, seeking support from genetic counselors, support groups, and research opportunities can provide valuable resources for coping with the challenges of abetalipoproteinemia.
\\
\hline
\end{tabular}
}
\caption{\textbf{Comparison of MedPhi.} Examples show qualitative differences in model generations with and without tuning vectors across medical tasks.}
\label{tab:appendix_qualitative_output_MEDphi}
\end{table*}

\subsection{System Prompt}
We use the following system prompt in benchmarking the accuracy of our models.
\begin{promptbox}[label={prompt:system}]{System prompt used.}
<|start\_header\_id|>system<|end\_header\_id|>
You are a highly qualified medical expert, adept at answering complex medical exam questions. Your task is to answer multiple-choice medical exam questions by following a structured reasoning process.
You will be given with some examples curated from a database of medical questions and their corresponding solutions. Each example showcases a detailed, step-by-step reasoning process leading to the correct answer. Carefully analyze these examples, paying close attention to the following:

    *   The medical knowledge applied in each step.
    
    *   The logical flow of reasoning, connecting medical concepts to the question.
    
    *   The format used to present the reasoning and final answer. 
    
Your task is to answer a new medical question by emulating the reasoning style and format observed in the examples.

Follow these instructions carefully:  
  
    1.  Thoroughly review the examples to internalize the reasoning patterns and solution format. While you should follow the format, DO NOT copy the reasoning verbatim. The idea is to understand how to THINK.
    
    2.  Read and understand the question.
    
    3.  Articulate your reasoning process inside the <think></think> tags. Connect each reasoning step to relevant medical concepts, demonstrating a deep understanding of the question's medical context.
    
    4.  Based on your reasoning, construct a well-structured explanation leading to the correct answer.
    
    5.  Finally, select the single correct answer option (A, B, C, or D).
    
    6.  Present your final answer in the format \"Answer: [X]\", where X represents the correct answer option you selected. It is crucial that you include NO additional text or tokens after ``Answer: [X]".
    
Here are the examples, presented in a user-assistant chat format, where the user poses a medical question, and the assistant provides a structured solution. The final user message will contain the new question you need to answer.

Question: The two principal contractile proteins found in skeletal muscle are:

Options:

A. actin and troponin.

B. actin and myosin.

C. troponin and tropomyosin.

D. myosin and tropomyosin.

<think>Actin is a thin filament protein, forming a helical structure along with tropomyosin and troponin, which regulate its interaction with myosin. Myosin, on the other hand, is a thick filament protein characterized by its long tail and globular head. The heads of the myosin filaments form cross-bridges by attaching to specific sites on the actin filaments. Through a series of movements fueled by ATP hydrolysis, these cross-bridges pull the actin filaments towards the center of the sarcomere, the basic unit of a muscle fiber, causing the muscle to contract. While troponin and tropomyosin are also important in muscle contraction, they serve more as regulatory proteins rather than primary contractile proteins.</think>

Answer: $[B]<|eot_id|>$
\end{promptbox}

\begin{table*}[!h]
\centering
\resizebox{\textwidth}{!}{
\begin{tabular}{l|cccccccccc|c}
\hline
\backslashbox{\textbf{Model}}{\textbf{Dataset}} & \textbf{BioRed} & \textbf{CareQA} & \textbf{Anata} & \textbf{Clin} & \textbf{Bio} & \textbf{Medi} & \textbf{Gene} & \textbf{Prof} & \textbf{MedMCQA} & \textbf{MedQA} & \textbf{Average} \\
\hline
\textbf{Qwen2.5-Aloe-Beta-7B} & \textbf{92.40} & \textbf{72.40} & \textbf{78.57} & \textbf{89.66} & \textbf{87.50} & \textbf{81.82} & \textbf{100.00} & \textbf{77.42} & \textbf{55.40} & \textbf{66.60} & \textbf{80.18} \\
\textbf{Qwen2.5-Math-Instruct} & 90.00 & 1.20 & 14.29 & 13.79 & 6.25 & 4.55 & 9.09 & 0.00 & 6.20 & 0.20 & 14.56 \\
\hline
\end{tabular}
}
\caption{\textbf{Performance on medical-domain becnhmarks.} Medically finetuning models perform better in medical tasks.(\textbf{Anata}: MMLU (anatomy), \textbf{Clin}: MMLU (clinical knowledge),\textbf{Bio}: MMLU (college biology), \textbf{Medi}: MMLU (college medicine), \textbf{Gene}: MMLU (medical genetics), and \textbf{Prof}: MMLU (professional medicine))}
\label{tab:ODD_med}
\end{table*}

\begin{table*}[!h]
\centering
\resizebox{\textwidth}{!}{
\begin{tabular}{l|ccccccccc|c}
\hline
\backslashbox{\textbf{Model}}{\textbf{Dataset}} & \textbf{MATH} & \textbf{Aqua Rat} & \textbf{Gaokao} & \textbf{GsM8k} & \textbf{Math qa} &\textbf{ Mine} & \textbf{coll} & \textbf{elem }& \textbf{high} & \textbf{Average} \\
\hline
\textbf{Qwen2.5-Aloe-Beta-7B} & 43.40 & 42.13 & 8.83 & 73.20 & 51.20 & 9.19 & \textbf{54.55} & 75.61 & 62.07 & 46.69 \\
\textbf{Qwen2.5-Math-Instruct} & \textbf{61.60} & \textbf{62.99} & \textbf{14.55} & \textbf{94.80} & \textbf{65.60} & \textbf{23.53} & \textbf{54.55} & \textbf{92.68} & \textbf{75.86} & \textbf{60.68} \\
\hline
\end{tabular}
}
\caption{\textbf{Performance on Math-domain becnhmarks.} Math finetuning models perform better in math benchmarks.(\textbf{Mine}: Minervamath, \textbf{coll}: MMLU (college mathematics), \textbf{elem}: MMLU (elementary mathematics), \textbf{high}: MMLU (high school mathematics))}
\label{tab:ODD_math}
\end{table*}

\begin{figure*}[!t]
\begin{center}
\includegraphics[width=0.8\textwidth]{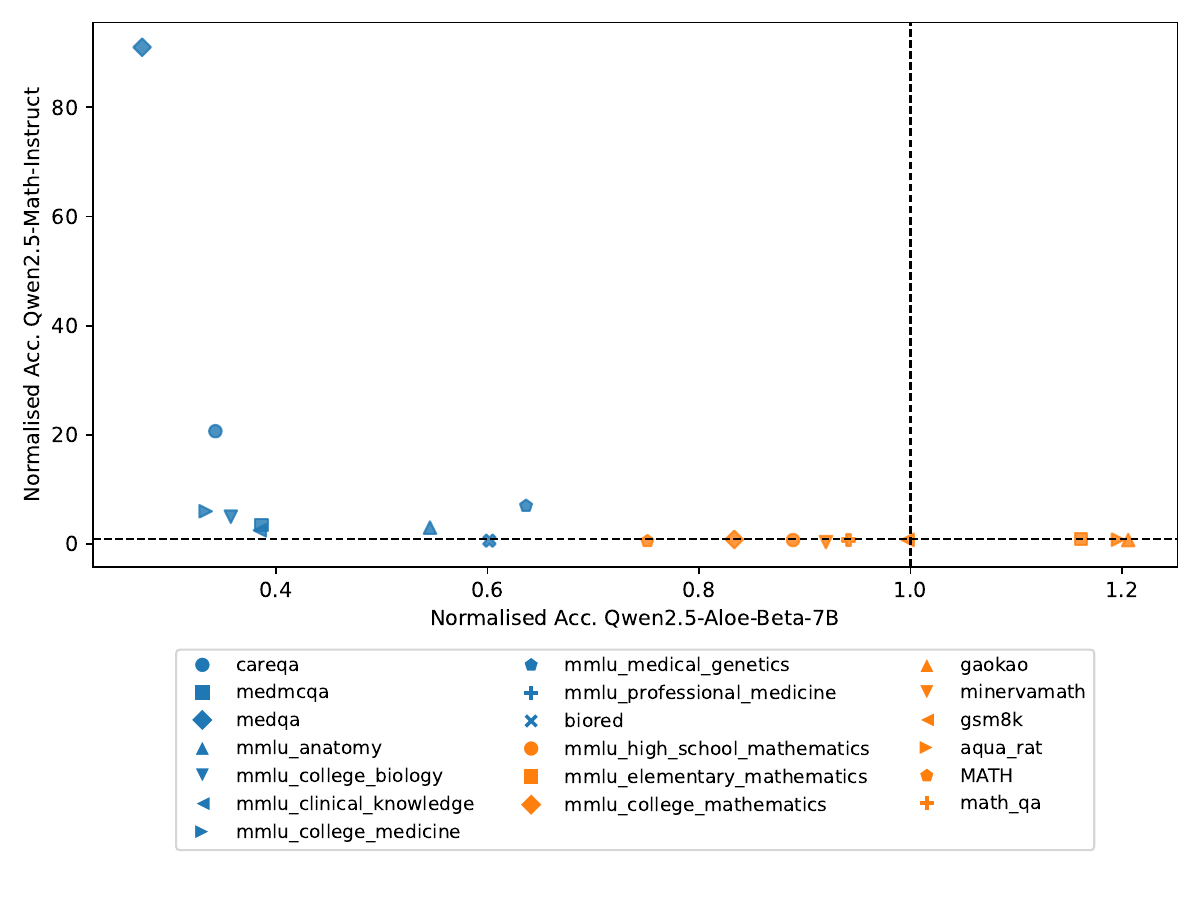}
\caption{The figure shows the normalised accuracy of the model created by adding qwen math and aloe vectors.}
\label{fig:adding_task_vector_med_math}
\end{center}
\vspace{-5mm}
\end{figure*}

\subsection{Attribute Patching}
\label{sec:attribution_patching_appndix}
Attribution patching~\cite{vig2020causal} is an interpretability technique used to identify which parts of a model (e.g., specific layers, heads, or MLP neurons) are causally responsible for a given behavior or prediction. The basic idea is to swap activations between two runs — one on a clean prompt (where the model behaves as desired) and another on a corrupted prompt (where a critical piece of information is removed or altered).
By measuring how the output changes when we “patch” activations from the clean run into the corrupted one, we can localize which internal components carry the relevant information. We run attribution patching on three variants of Meta-LLaMA-3-7B: base, instruct, and aloe, using the following clean and corrupt prompts:

\begin{tcolorbox}[colback=gray!5, colframe=gray!40, title=Clean vs. Corrupt Prompts, boxrule=0.4pt, arc=2mm]
\textbf{Clean Prompt:}\\
\texttt{"Prednisolone can treat eye inflammation. This statement is:"}

\vspace{3mm}
\textbf{Corrupt Prompt:}\\
\texttt{"Clofoctol can treat eye inflammation. This statement is:"}
\end{tcolorbox}

\begin{figure*}[!ht]
\begin{center}
\includegraphics[width=0.8\textwidth]{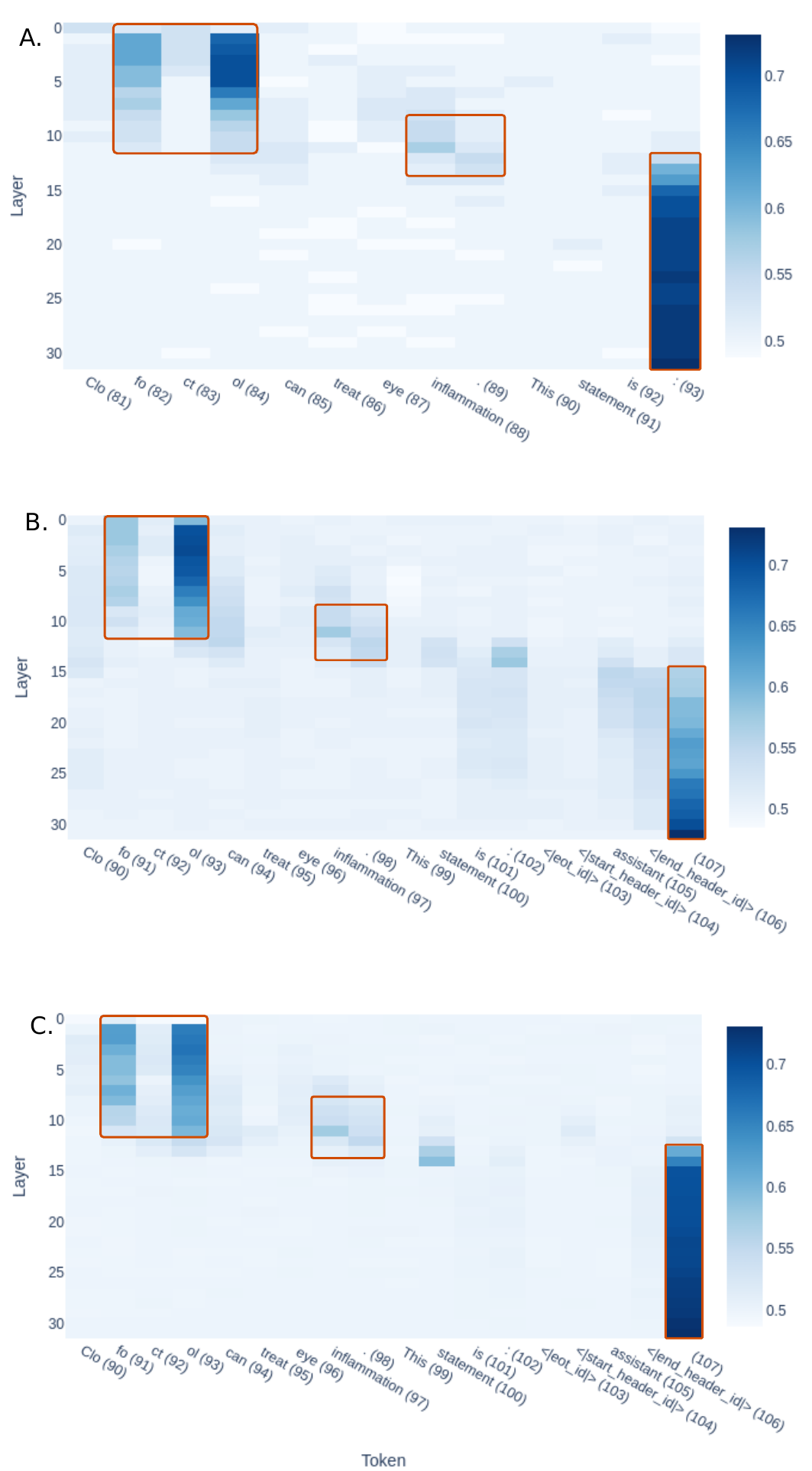}

\caption{We find similar regions of interest at similar layer and token positions across all Meta-LLaMA-3-7B variants: A) Base, B)Instruct, and C) Aloe.(ref. Appendix~\ref{sec:attribution_patching_appndix} for a detailed discussion)}
\label{fig:attribute_patching}
\end{center}
\vspace{-5mm}
\end{figure*}

Our prompts are based on understanding if the models can access factually accurate statements of now. As seen in Table~\ref{fig:attribute_patching}, we observe similar activation regions across all three models. We believe these regions correspond to analogous functional stages: processing the input, interpreting the query, and generating the final answer~\cite{marks2023geometry}. For the Instruct and Aloe variants, we also identify an additional region near the end-of-input (EOI) token, which may be attributed to differences in their chat template training setups.

\subsection{Value of k}
\label{sec:value_of_k}

The number of singular vectors used for our analysis, $k$, is determined by minimising the approximation error,$\epsilon$ to 0.05. This can be expressed in terms of the singular values $\sigma_i$ of $W$ as

\[
k = \min \left\{ k : \frac{\sum_{i=k+1}^{n} \sigma_i^2}{\sum_{i=1}^{n} \sigma_i^2} \le \varepsilon^2 \right\}.
\]

\subsection{Cosine similarity between weights of pretrained and fine-tuned models.}

Table~\ref{tab:model_weight_similarity} shows the similarity between weights of models
\begin{table}[h!]
\centering
\begin{adjustbox}{width=0.55\linewidth}
\begin{tabular}{c|c|c}
    \textbf{Model} & \multicolumn{2}{c}{\textbf{Similarity}} \\
    \hline
    \multicolumn{3}{c}{\cellcolor[HTML]{EEEEEE}\textbf{Meta-Llama-3-B}} \\
    \hline
    & \textbf{Instruct} & \textbf{Aloe $\boldsymbol{\alpha}$} \\
    \hline
    \textbf{Base} & \cellcolor{blue!20}99.93 & \cellcolor{blue!20}99.75 \\
    \hline
    \multicolumn{3}{c}{\cellcolor[HTML]{EEEEEE}\textbf{Qwen 2.5-7B}} \\
    \hline
    & \textbf{Instruct} & \textbf{Aloe $\boldsymbol{\alpha}$} \\
    \hline
    \textbf{Base} & \cellcolor{blue!20}99.99 & \cellcolor{blue!20}99.98 \\
    \hline
    \multicolumn{3}{c}{\cellcolor[HTML]{EEEEEE}\textbf{Phi-3.5}} \\
    \hline
    & \multicolumn{2}{c}{\textbf{MedPhi}} \\
    \hline
    \textbf{Instruct} & \multicolumn{2}{c}{\cellcolor{blue!20}99.98} \\
\end{tabular}
\end{adjustbox}
\caption{\textbf{Cosine similarity between weights of pre-trained and fine-tuned models (scores $\times 100$).} The consistently high similarity score indicates that fine-tuning only marginally alters the underlying representation space.}
\label{tab:model_weight_similarity}
\end{table}

\subsection{Cross-domain performance of domain-specific models.}
We evaluate the robustness of the specialised model in performing out-of-distribution tasks. We evaluate the performance of Qwen2.5-Aloe-beta-7B and Qwen2.5-Math-Instruct in Medical and Math benchmarks. Tables~\ref{tab:ODD_math} and ~\ref{tab:ODD_med} show the performance if the two models in Math and Medical benchmarks, respectively. We note that models fail to perform well on out-of-distribution tasks

\end{document}